\documentclass[sigconf]{acmart}

\usepackage{subcaption}
\usepackage{subfiles}
\usepackage{fontawesome}
\usepackage{multirow}
\usepackage{tabularx} %
\usepackage{makecell} %

\usepackage{amsmath,amsfonts,bm}

\def\eqref#1{equation~\ref{#1}}

\def\1{\bm{1}}

\DeclareMathAlphabet{\mathsfit}{\encodingdefault}{\sfdefault}{m}{sl}
\SetMathAlphabet{\mathsfit}{bold}{\encodingdefault}{\sfdefault}{bx}{n}

\def\gG{{\mathcal{G}}}

\newcommand{\benchmark}{\textsc{$\mathcal{H}^2$GB}}
\newcommand{\framework}{{\textsc{UnifiedGT}}}
\newcommand{\model}{$\mathcal{H}^2$G-former}
\newcommand{\dataset}[1]{\texttt{#1}}
\newcommand{\hindex}{$\mathcal{H}^2$}

\theoremstyle{definition}
\newtheorem{definition}{Definition}

\makeatletter
\def\thickhline{%
  \noalign{\ifnum0=`}\fi\hrule \@height \thickarrayrulewidth \futurelet
   \reserved@a\@xthickhline}
\def\@xthickhline{\ifx\reserved@a\thickhline
               \vskip\doublerulesep
               \vskip-\thickarrayrulewidth
             \fi
      \ifnum0=`{\fi}}
\makeatother

\newlength{\thickarrayrulewidth}
\usepackage{enumitem,cleveref,amsmath}
\usepackage{soul}

\AtBeginDocument{%
  \providecommand\BibTeX{{%
    \normalfont B\kern-0.5em{\scshape i\kern-0.25em b}\kern-0.8em\TeX}}}

\copyrightyear{2025}
\acmYear{2025}
\setcopyright{cc}
\setcctype{by}
\acmConference[KDD '25]{Proceedings of the 31st ACM SIGKDD Conference on Knowledge Discovery and Data Mining V.2}{August 3--7, 2025}{Toronto, ON, Canada}
\acmBooktitle{Proceedings of the 31st ACM SIGKDD Conference on Knowledge Discovery and Data Mining V.2 (KDD '25), August 3--7, 2025, Toronto, ON, Canada}
\acmDOI{10.1145/3711896.3737421}
\acmISBN{979-8-4007-1454-2/2025/08}

\begin{document}

\title{When Heterophily Meets Heterogeneity: \\ Challenges and a New Large-Scale Graph Benchmark}

\author{Junhong Lin}
\email{junhong@mit.edu}
\affiliation{%
  \institution{Massachusetts Institute of Technology}
  \city{Cambridge}
  \state{MA}
  \country{United States}
}

\author{Xiaojie Guo}
\email{Xiaojie.Guo@ibm.com}
\affiliation{%
  \institution{IBM Research}
  \city{Yorktown Heights}
  \state{NY}
  \country{United States}
}

\author{Shuaicheng Zhang}
\email{zshuai8@vt.edu}
\affiliation{%
  \institution{Virginia Tech}
  \city{Blacksburg}
  \state{VA}
  \country{United States}
}

\author{Yada Zhu}
\email{yzhu@us.ibm.com}
\affiliation{%
  \institution{IBM Research}
  \city{Yorktown Heights}
  \state{NY}
  \country{United States}
}

\author{Julian Shun}
\email{jshun@mit.edu}
\affiliation{%
  \institution{Massachusetts Institute of Technology}
  \city{Cambridge}
  \state{MA}
  \country{United States}
}

\begin{abstract}
Graph mining has become crucial in fields such as social science, finance, and cybersecurity. Many large-scale real-world networks exhibit both heterogeneity, where multiple node and edge types exist in the graph, and heterophily, where connected nodes may have dissimilar labels and attributes. 
However, existing benchmarks primarily focus on either heterophilic homogeneous graphs or homophilic heterogeneous graphs, leaving a significant gap in understanding how models perform on graphs with both heterogeneity and heterophily.
To bridge this gap, we introduce \benchmark{}, a large-scale node-classification graph benchmark that brings together the complexities of both the heterophily and heterogeneity properties of real-world graphs. \benchmark{} encompasses 9 real-world datasets spanning 5 diverse domains, 28 baseline models, and a unified benchmarking library with a standardized data loader, evaluator, unified modeling framework, and an extensible framework for reproducibility. We establish a standardized workflow supporting both model selection and development, enabling researchers to easily benchmark graph learning methods.
Extensive experiments across 28 baselines reveal that current methods struggle with heterophilic and heterogeneous graphs, underscoring the need for improved approaches. Finally, we present a new variant of the model, \model{}, developed following our standardized workflow, that excels at this challenging benchmark.
Both the benchmark and the framework are publicly available at \href{https://github.com/junhongmit/H2GB}{\textcolor{blue}{Github}} and \href{https://pypi.org/project/H2GB}{\textcolor{blue}{PyPI}}, with documentation hosted at \href{https://junhongmit.github.io/H2GB}{\textcolor{blue}{https://junhongmit.github.io/H2GB}}. 
\end{abstract}



\maketitle

\section{Introduction}
\label{sec:introduction}
\begin{figure*}[t!]
    \centering
    \includegraphics[width=0.8\textwidth]{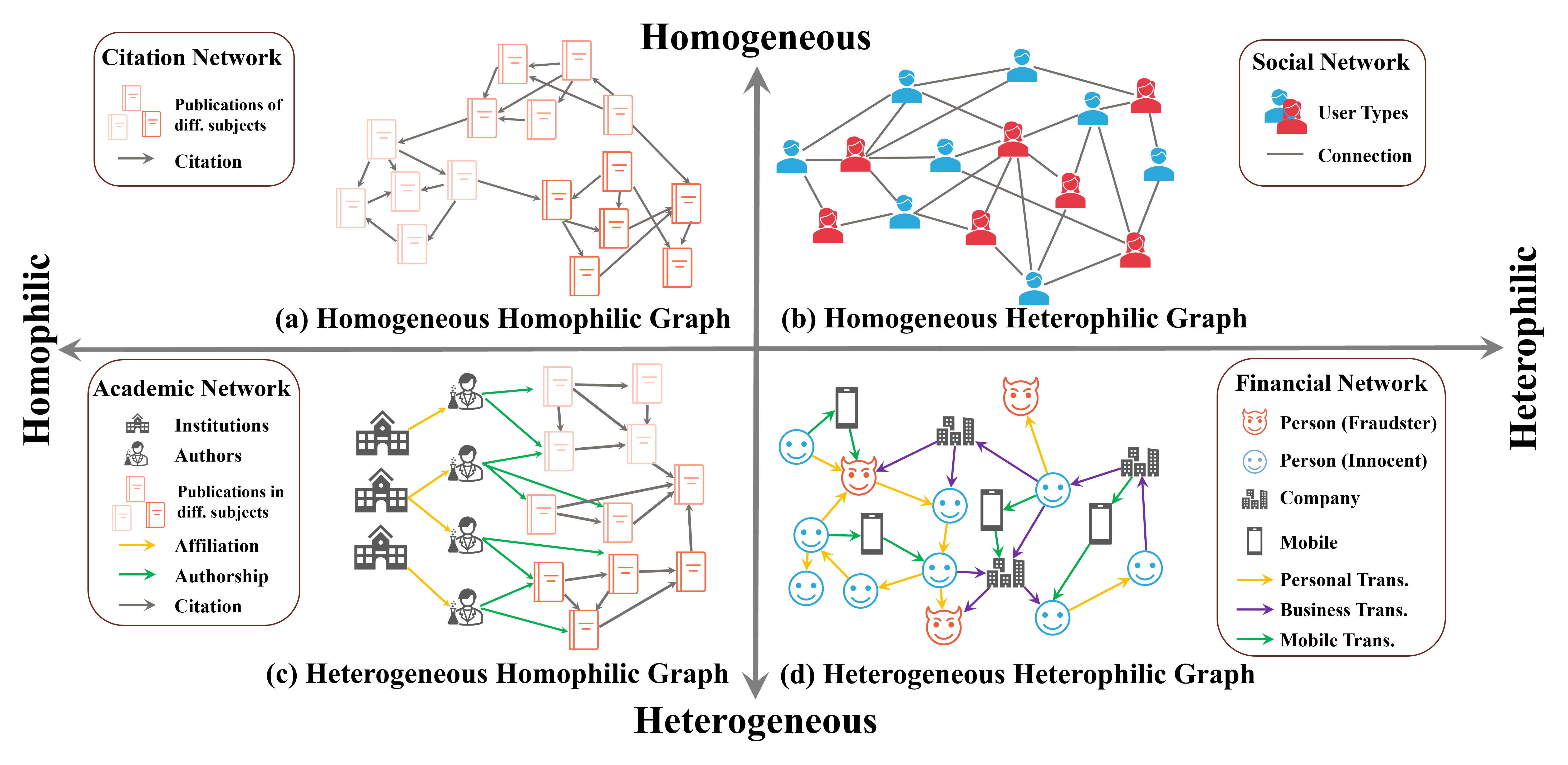}
    \caption{\small Examples of graphs with different levels of heterophily and heterogeneity. Nodes with different class labels and edges of different types are represented with different colors (e.g., publications in different subjects or different kinds of financial transactions).}
    \label{fig:example}
\end{figure*}

Graphs are commonly used to model complex relationships across various domains, such as finance~\cite{wang2019semi}, social science~\citep{takac2012data, leskovec2012learning} and cybersecurity~\cite{he2022illuminati, warmsley2022survey}. Many real-world graphs contain millions or even billions of nodes and edges, making scalable learning methods essential.
Graph neural networks (GNNs) \citep{hamilton2017inductive, kipf2016semi} have achieved state-of-the-art performance on graph learning tasks. 
However, they were designed primarily for \textit{homogeneous homophilic graphs}, where the nodes and edges are of a single type~\citep{gilmer2017neural, zhu2020beyond}, and connected nodes are similar, as shown in \Cref{fig:example}(a).

As real-world graphs grow in scale, they increasingly exhibit heterogeneity and heterophily. \textit{\textbf{Heterogeneity}} arises from multiple entity and relation types, adding structural and semantic complexity. This diversity, in turn, intensifies \textit{\textbf{heterophily}}, the tendency for connected nodes to have dissimilar labels or attributes. For example, financial networks (\Cref{fig:example}(d))~\citep{rao2021xfraud, altman2024realistic} contain diverse node types (e.g., person, business)  and edge types (e.g., wire transfer, check transaction). Furthermore, fraudsters tend to have different labels than their innocent neighbors, making these networks both heterogeneous and heterophilic. These properties, common in domains such as e-commerce~\citep{liu2023datasets}, academia~\citep{zhang2019oag, hu2021ogb}, and cybersecurity~\citep{kumarasinghe2022pdns, aravind2022malicious}, pose significant challenges to GNN performance.

In recent years, researchers have actively explored methods to overcome these challenges in two separate directions.
First, to handle graphs with heterophily, there has been a recent line of research on developing heterophilic graph benchmarks~\citep{bo2021beyond, lim2021large} and heterophily-centered GNNs~\citep{zhu2020beyond, luan2022revisiting, pei2020geom, zhu2021graph, bo2021beyond} that incorporate long-range relationships and distinct aggregation mechanisms, such as distant node exploration~\citep{zhu2020beyond, pei2020geom, abu2019mixhop, li2022finding}, signed aggregation~\citep{bo2021beyond, zhu2021graph, luan2022revisiting}, and local grouping~\citep{li2022finding}. However, these heterophilic GNNs are restricted to homogeneous graphs, as illustrated in \Cref{fig:example}(b). Second, heterogeneous GNNs have been proposed to handle the diverse information present in heterogeneous graphs~\citep{schlichtkrull2018modeling, zhang2019heterogeneous, wang2019heterogeneous, fu2020magnn, hong2020attention, hu2020heterogeneous}. However, most heterogeneous GNNs are implicitly built upon the homophily assumption, as illustrated in \Cref{fig:example}(c), and exhibit poor performance on heterophilic graphs~\citep{guo2023homophily}.

While there has been recent progress on handling heterogeneity and heterophily separately, many large real-world graphs exhibit both properties simultaneously. 
A recent research effort, the \textit{Heterophily Graph Learning Handbook}~\citep{luan2024heterophilic}, explicitly highlights this gap, emphasizing that previous research primarily evaluated models on graphs that focused on either only heterophily or only heterogeneity.
The following challenges arise when exploring graph learning in heterophilic and heterogeneous settings. \textbf{(1) Lack of benchmarks for graphs with both heterophily and heterogeneity~\citep{luan2024heterophilic}}: Existing benchmarks either focus exclusively on homogeneous graphs, neglecting the diversity of node and edge types found in real-world graphs, or on heterogeneous graphs while assuming homophily.
\textbf{(2) Limited understanding of heterophily in heterogeneous Graphs~\citep{luan2024heterophilic}}:
Heterophily has been largely studied in homogeneous graphs, leaving its impact on heterogeneous structures under-explored. This gap limits our understanding of how heterophilic patterns interact with diverse node and edge types. \citet{guo2023homophily} found that heterogeneous GNNs often degrade in performance under heterophily, highlighting the need for better modeling strategies.
\textbf{(3) Inadequacy of heterophilic GNNs on large-scale heterogeneous graphs}:
Heterophilic GNNs are typically designed for homogeneous graphs, making them ineffective in heterogeneous settings where node and edge types vary. They also struggle to scale with graph size, as many were developed for small graphs, limiting their applicability to large real-world networks.

\begin{figure*}[t!]
    \centering
    \includegraphics[width=0.95\textwidth]{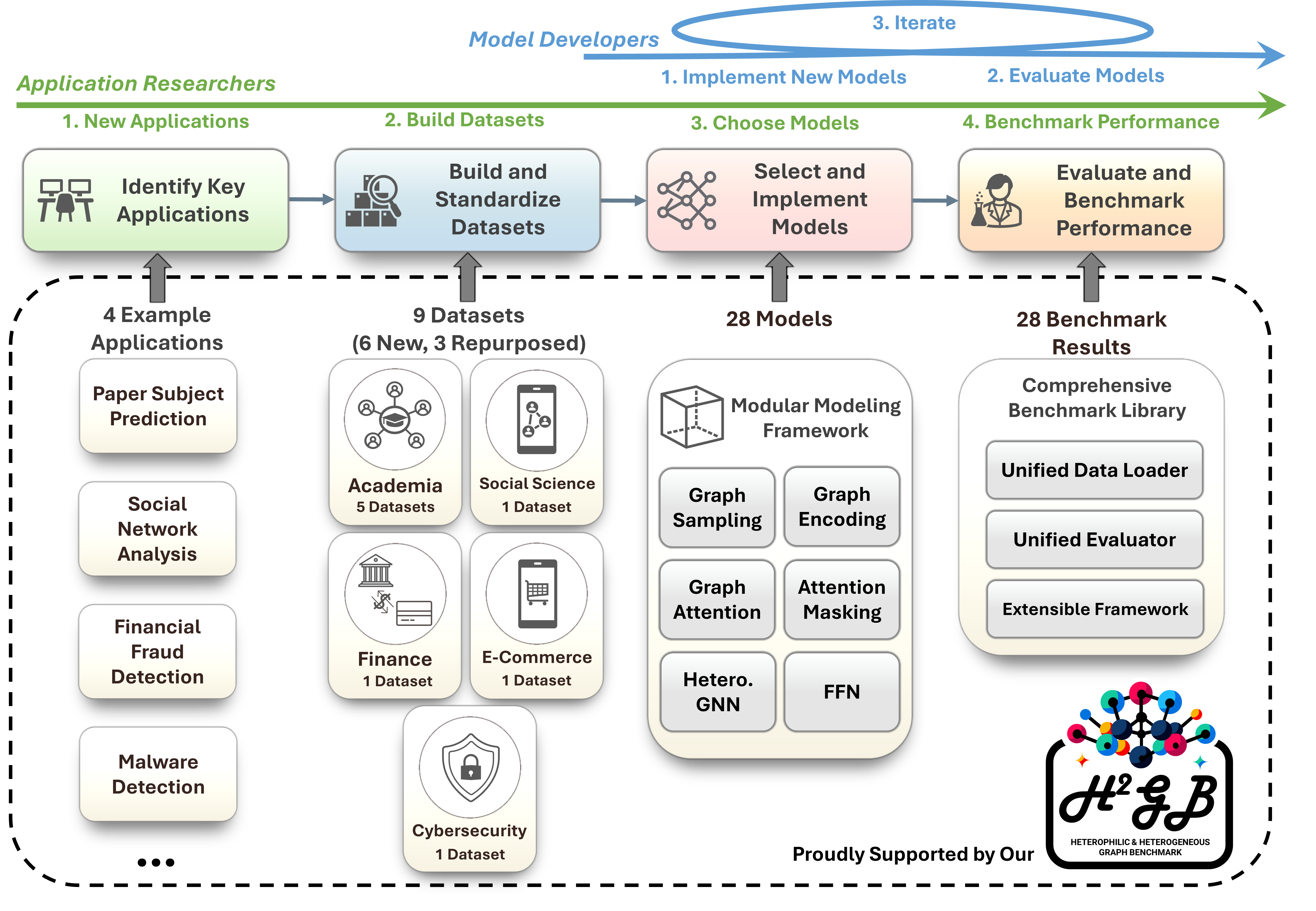}
    \caption{\small \benchmark{} offers a complete benchmark workflow for heterophilic and heterogeneous graph learning, featuring a diverse dataset suite (\Cref{sec:datasets}), a modular modeling framework (\Cref{sec:framework}), and a comprehensive benchmark library, making it easy to evaluate and compare different methods (\Cref{sec:experiments}). The green and blue arrows on top highlight two standard workflows for users to interact with \benchmark{}.
    }
    \label{fig:benchmark}
\end{figure*}

To address these challenges, we introduce the \underline{\textbf{H}}eterophilic and \underline{\textbf{H}}eterogeneous \underline{\textbf{G}}raph \underline{\textbf{B}}enchmark (\benchmark{}), the first, novel and comprehensive graph benchmark designed to evaluate graph learning methods on large-scale heterophilic and heterogeneous graphs across multiple real-world domains. As shown in \Cref{fig:benchmark}, \benchmark{} provides the following contributions:
\begin{itemize}[topsep=1pt, leftmargin=15pt]
\item \textbf{Diverse Real-World Datasets}: \benchmark{} consists of 4 applications, and 9 real-world datasets spanning 5 domains: academia, finance, e-commerce, social science, and cybersecurity.

\item \textbf{Standardized Benchmarking}: \benchmark{} establishes a standardized evaluation framework for node classification, providing an extensive comparison of \textit{28 baseline models} implemented through our previously built modular graph learning framework, \framework{}~\citep{lin2024unifiedgt}, including message-passing GNNs, graph transformers, and non-GNN baselines, under a unified experimental setup.

\item \textbf{Standardized Workflow}: We introduce a standard workflow supporting both model selection and development. In particular, we demonstrate a case study using \benchmark{} for the development of a new model in \Cref{sec:case_study}.

\item \textbf{New Heterophily Measure}: Existing metrics (e.g., edge heterophily) provide limited insights into heterogeneous graph structures. We introduce a \textit{new heterophily measure, the \hindex{} index,} which better captures complex heterophilic interactions, addressing a key limitation identified in prior literature~\cite{luan2024heterophilic}.

\item \textbf{Scalability Focus}: \benchmark{} emphasizes scalability by evaluating graph learning methods on large-scale heterophilic and heterogeneous graphs. Most of our datasets are large, containing millions of nodes and tens of millions of edges (see \Cref{table:stats}), which are orders of magnitude larger than existing heterophilic benchmarks~\citep{zheng2022graph, lim2021large}. \benchmark{} evaluates models across large-scale graphs, identifies the performance bottlenecks of existing GNNs, and encourages the development of scalable heterophilic graph learning methods.

\item \textbf{Open-Source Benchmarking Library}: \benchmark{} is released as an extensible and user-friendly Python library consisting of a unified data loader and evaluator, making it easy to access datasets, evaluate methods, and compare performance.
\end{itemize}

Through comprehensive experiments on our datasets, we draw the following insights: (1) homogeneous heterophilic GNNs underperform heterogeneous homophilic GNNs due to their inability to account for diverse node and edge types;
(2) performance varies significantly among heterogeneous homophilic GNNs, likely due to differences in their architectural robustness when exposed to heterophily; and
(3) non-scalable GNNs struggle on our large-scale heterogeneous heterophilic benchmark.
Lastly, following our established standard workflow, we develop \model{}, a new effective model variant by incorporating several new components including masked label embedding, heterogeneous attention, $k$-hop attention mask, and type-specific FFNs, significantly improving performance on datasets in \benchmark{}.

\begin{table*}[!t]
\caption{\small Statistics of \benchmark{} datasets. \#C is the number of classes, with imbalance ratios provided for binary classification. The training/validation/test split ratio is indicated under the Split Scheme.}
\label{table:stats}
\begin{center}
\resizebox{0.85\textwidth}{!}{%
\begin{tabularx}{0.93\textwidth}{@{}c@{\hspace{0.4em}}r@{\hspace{0.4em}}l@{\hspace{0.4em}}r@{\hspace{0.4em}}l@{\hspace{0.4em}}rc@{\hspace{0.4em}}c@{\hspace{0.4em}}c@{\hspace{0.4em}}c@{\hspace{0.4em}}cc@{}}
\toprule
 Dataset & \# Nodes & (types) & \# Edges & (types) & \# Feat. & \# C (Ratio) & Label & Split Scheme (Ratio [\%]) & Metric\\
\midrule
\dataset{ogbn-mag} & 1,939,743 & (4) & 42,182,144 & (7) & 128 & 349  & paper venue & Time (85/9/6) & Accuracy\\
\dataset{mag-year} & 1,939,743 & (4) & 42,182,144 & (7) & 128 & 5 & publication year & Random (50/25/25) & Accuracy\\
\dataset{oag-cs} & 1,112,691 & (4) & 27,537,448 & (22) & 768 & 3,514 & paper venue & Time (80/9/11) & Accuracy\\
\dataset{oag-eng} & 929,315 & (4) & 12,346,854 & (22) & 768 & 3,956 & paper venue & Time (88/10/2) & Accuracy\\
\dataset{oag-chem} & 1,918,881 & (4) & 38,098,014 & (22) & 768 & 2,985 & paper venue & Time (90/8/2) & Accuracy\\
\dataset{RCDD} & 13,806,619 & (7) & 157,814,864 & (14) & 256 & 2 (11:1) & risk commodity & Time (70/15/15) & F1 score\\
\dataset{IEEE-CIS-G} & 153,880 & (12) & 2,873,472 & (22) & 4823 & 2 (12:1) & fraud transaction & Time (80/10/10) & F1 score\\
\dataset{H-Pokec} & 1,731,977 & (16) & 51,774,836 & (31) & 66 & 2 (1:1) & gender & Random (50/25/25) & Accuracy\\
\dataset{PDNS} & 1,173,558 & (2) & 76,797,104 & (4) & 10 & 2 (1:2) & malicious domain & Time (70/20/10) & F1 score\\
\bottomrule
\end{tabularx}
}
\end{center}
\end{table*}

\section{Preliminaries and Related Work}
\label{sec:related_works}
\begin{definition}[Graph Heterogeneity]
A heterogeneous graph is a directed graph $\gG=(\mathcal{V}, \mathcal{E}, \mathcal{A}, \mathcal{R})$, where each node $v\in\mathcal{V}$ and edge $e\in\mathcal{E}$ has a type given by $\tau(v): V\rightarrow\mathcal{A}$ and $\phi(e):E\rightarrow\mathcal{R}$. Here, $\mathcal{A}$ and $\mathcal{R}$ are the set of node and edge types, respectively.
\end{definition}

\begin{definition}[Metapath-Induced Subgraphs] A metapath is a sequence of edges, defined as $\mathcal{P} = A_1 \xrightarrow{R_1} A_2 \xrightarrow{R_2} \cdots A_{n}\xrightarrow{R_n} A_{n+1}$, where $A_i \in \mathcal{A}$ and $R_i \in \mathcal{R}$. Given a metapath $\mathcal{P}$, we can construct a metapath-induced subgraph $\mathcal{G_P}$, which includes edge $(u, v)$ in $\mathcal{G_P}$ 
if and only if there exists at least one length-$n$ path between $u$ and $v$ following the metapath $\mathcal{P}$ in the original graph $\mathcal{G}$.
\end{definition}

\begin{definition}[Graph Heterophily]
Graph heterophily quantifies the dissimilarity between connected nodes based on their attributes or labels. Common metrics such as edge heterophily~\citep{zhu2020beyond} and node heterophily~\citep{pei2020geom} are designed for homogeneous graphs, quantifying the proportion of connected nodes that have different labels.
\end{definition}

\begin{definition}[Node Classification Task]
Given a graph $\mathcal{G} = (\mathcal{V}, \mathcal{E}, \mathcal{A}, \mathcal{R})$, only a subset of nodes of a specific type $\mathcal{V}_T \subseteq \mathcal{V}$ (\textit{task entities}) are labeled. The task is to learn a function $f:(\mathcal{G}, v) \mapsto y_v$ that predicts the label $y_v$ for unlabeled nodes $v \in \mathcal{V}_T$.
\end{definition}

\subsubsection*{\textbf{Graph Learning for Heterogeneous and Heterophilic Graphs.}} Existing heterogeneous GNNs are classified into \textit{metapath-based} methods, which extract structural information from homogeneously-typed subgraphs by predefined metapaths to capture diverse semantic data \citep{schlichtkrull2018modeling, zhang2019heterogeneous, wang2019heterogeneous, fu2020magnn}, and \textit{metapath-free} methods, which process structural and semantic information simultaneously, enhancing message aggregation by incorporating node and edge types without relying on predefined paths \citep{zhu2019relation, hong2020attention, hu2020heterogeneous, lv2021we}. While these approaches take heterogeneity into account, they generally maintain the homophily assumption. In contrast, existing heterophilic GNNs have been tailored primarily for homogeneous graphs and lack mechanisms to address heterogeneity \citep{abu2019mixhop, bo2021beyond, lim2021large}. Recent works aim to bridge this gap by improving heterophilic learning on heterogeneous graphs through augmented graphs and disentangled loss functions \citep{guo2023homophily, li2023hetero}; however, they primarily focus on enhancing existing models rather than introducing fundamentally new solutions optimized for both heterophily and heterogeneity.

\subsubsection*{\textbf{Current Datasets.}} Recent evaluations of heterophilic graph learning primarily use small-scale datasets from Pei et al.~\citep{pei2020geom}. Lim et al.~\citep{lim2021large} have compiled larger non-homophilic 
graph datasets, which have become the standard for evaluating heterophilic GNNs, but their datasets are limited to homogeneous graphs. Several heterogeneous academic network datasets have been introduced, including \dataset{DBLP}~\citep{lv2021we}, \dataset{ACM}~\citep{lv2021we}, \dataset{ogbn-mag}~\citep{hu2020open}, \dataset{MAG240M}~\citep{hu2021ogb}, and \dataset{IGB}~\citep{igbdatasets}. 
However, these datasets have not been tested with heterophilic GNN methods. 
Moreover, the pure focus on academic networks narrows their use in addressing graph learning challenges in other domains.

\subsubsection*{\textbf{Conventional Heterophily Metrics}}\label{sec:typical_heterophily}

Typical heterophily metrics, such as edge heterophily ($\mathcal{H}_{\text{edge}}$)~\citep{zhu2020beyond}, node heterophily ($\mathcal{H}_{\text{node}}$)~\citep{pei2020geom}, and adjusted heterophily ($\mathcal{H}_{\text{adj}}$)~\citep{platonov2024characterizing}, are designed for homogeneous graphs, quantifying different aspects of label mixing among connected nodes. While edge and node heterophily directly reflect label differences along edges or within local neighborhoods, they are sensitive to class imbalance~\citep{lim2021large}. Adjusted heterophily mitigates this issue by normalizing based on class distributions. A common approach to extend these metrics to heterogeneous graphs is to disregard node and edge types, treating the graph as homogeneous. Yet, this simplification overlooks structural dependencies across different node types. Traditional metrics typically assess heterophily only among nodes of the same type, failing to account for homophily that may emerge along metapath-based structures.
\citet{guo2023homophily} empirically showed that heterogeneous GNNs perform better when metapath-induced subgraphs are homophilic, a factor not captured by typical heterophily measures. Consequently, these metrics can misrepresent a model’s true ability to handle heterophilic relationships in heterogeneous graphs.

This limitation underscores the need for a better heterophily measure designed for heterogeneous graphs.
Recent works~\citep{liu2023datasets, guo2023homophily} have proposed the metapath-based label heterophily (MLH) measure, which extends edge heterophily, $\mathcal{H}_{\text{edge}}$, to a metapath-induced subgraph $\mathcal{G_P}$, and is formulated as follows:  
\begin{align}\label{eq:metapath_edge_heterophily}
\text{MLH}(\mathcal{G}) &= \text{Agg}(\mathcal{H}_{\text{edge}}(\mathcal{G_P}) | \mathcal{P}\in \mathcal{M}_k),
\end{align}
where $\mathcal{M}_k$ denotes a $k$-hop metapath set, and $\text{Agg} \in \{\text{mean}, \text{max}\}$.
However, it suffers from class imbalance~\citep{lim2021large}, leading to artificially low values (indicating homophily) in datasets that are inherently heterophilic. For instance, as shown in \Cref{table:typical_heterophily}, the \dataset{RCDD} and \dataset{IEEE-CIS-G} datasets demonstrate significant class imbalance, which contributes to deceptively low MLH values.

\section{Heterophilic and Heterogeneous Graph Benchmark (\benchmark{})} 
\label{sec:datasets}
In this section, we present \benchmark{}, a benchmark consisting of 9 large-scale datasets (6 new ones and 3 from existing work), shown in \Cref{table:stats}, spanning 5 diverse domains (\Cref{fig:benchmark}): academia, e-commerce, finance, social science, and cybersecurity. We also introduce a new heterophily measure that better captures the heterophilic properties of heterogeneous graphs.
The benchmark standardizes data loading, data splitting, feature encoding, and performance evaluation, which together enable open and reproducible research on heterophilic and heterogeneous GNNs.\footnote{As a special case, \benchmark{} can also be useful for systematically evaluating homogeneous GNNs (by simply applying a learnable type-dependent feature projection and then ignoring the type information on nodes and edges).}

\begin{table*}[!t]
\caption{Heterophily measures on each dataset. A value near 0 indicates homophily, where nodes primarily connect to others of the same class, while values around 1 suggest heterophily, where nodes prefer connections to different classes. $y_v$ is the label of node $v$, $C$ denotes the number of classes, $d(v)$ is the in-degree of node $v$, $D_k=\sum_{v:y_v=k}{d(v)}$ is the total in-degree of class $k$ nodes, $\mathcal{M}_k$ is a $k$-hop metapath set, and $\text{Agg} \in \{\text{mean}, \text{max}\}$ is an aggregation function.}
\label{table:typical_heterophily}
\begin{center}
\resizebox{\textwidth}{!}{%
\newcolumntype{R}{>{}r<{}}
\newcolumntype{L}{>{}l<{}}
\newcolumntype{M}{R@{${}={}$}L}
\begin{tabularx}{1.24\textwidth}{cM@{\hspace{0.8em}}c@{\hspace{0.8em}}c@{\hspace{0.8em}}c@{\hspace{0.8em}}c@{\hspace{0.8em}}c@{\hspace{0.8em}}c@{\hspace{0.8em}}c@{\hspace{0.8em}}c@{\hspace{0.8em}}c}
\toprule
\multicolumn{3}{c}{Heterophily Metric}
& \dataset{ogbn-mag} & \dataset{mag-year} & \dataset{oag-cs} & \dataset{oag-eng} & \dataset{oag-chem} & \dataset{RCDD} & \dataset{IEEE-CIS-G} & \dataset{H-Pokec} & \dataset{PDNS} \\
\midrule
Edge Heterophily  & $\displaystyle \mathcal{H}_{\text{edge}}$ & $\displaystyle \frac{|\{(u,v) \in \mathcal{E} : y_u \ne y_v\}|}{|\mathcal{E}|}$ & 0.9205 & 0.7909 & 0.9835 & 0.9586 & 0.9457 & 0.5001 & 0.5917 & 0.5663 & 0.4990 \\
\midrule
Node Heterophily  & $\displaystyle \mathcal{H}_{\text{node}}$ & $\displaystyle \frac{1}{|\mathcal{V}|} \sum_{v \in \mathcal{V}} \frac{ | \{u \in \mathcal{N}(v): y_v \ne y_u \} |  } { |\mathcal{N}(v)| }$ & 0.9539 & 0.7946 & 0.9880 & 0.9748 & 0.9696 & 0.5005 & 0.5839 & 0.5667 & 0.4992 \\
\midrule
Adjusted Heterophily & $\displaystyle \mathcal{H}_{\text{adj}}$ & $\displaystyle 1 - 
\frac{1-\sum_{k=1}^{C}{D_k^2/(2|\mathcal{E}|)^2}-\mathcal{H}_{\text{edge}}}{1-\sum_{k=1}^{C}{D_k^2/(2|\mathcal{E}|)^2}} $ & 0.9312 & 0.9977 & 0.9847 & 0.9612 & 0.9496 & 0.8398 & 1.3151 & 1.1350 & 1.0027 \\
\midrule
\thead{Metapath-based Label Heterophily} & MLH & $\text{Agg}\left(\mathcal{H}_{\text{edge}}(\mathcal{G_P})\left| \mathcal{P}\in \mathcal{M}_k\right.\right)$ & 0.8731 & 0.7718 & 0.9623 & 0.8689 & 0.8724 & 0.4912 & 0.1352 & 0.3922 & 0.3916 \\
\midrule
\textbf{\hindex{} Index (Ours)} & $\displaystyle \mathcal{H}^2$ & $ \text{Agg}\left(\mathcal{H}_{\text{adj}}(\mathcal{G_P}) \left| \mathcal{P}\in \mathcal{M}_k \right.\right)$ & 0.8773 & 0.9654 & 0.9652 & 0.8729 & 0.8858 & 0.9776 & 0.9846 & 0.9488 & 0.7866 \\
\bottomrule
\end{tabularx}
}
\end{center}
\end{table*}

\subsection{Key Applications}
Real-world graphs exhibit a diverse range of applications, many of which inherently involve both heterophily and heterogeneity. We identify four representative key real-world applications where such graph structures naturally arise: paper venue classification, social network analysis, financial fraud detection, and malware detection. They span diverse domains---academia, finance, e-commerce, cybersecurity, and social science---each presenting unique challenges that demand robust graph learning methods, ensuring that \benchmark{} captures the complexities of large-scale real-world heterophilic and heterogeneous graphs across multiple domains.

\subsubsection*{\textbf{Paper Venue Classification.}} In academic networks, papers are often connected through citations, co-authorships, or shared topics. While prior studies typically assume a homophilic structure where related papers belong to the same venue, real-world academic graphs exhibit heterophily---papers from the same author often span multiple venues and disciplines. We study this using one existing dataset, \dataset{ogbn-mag}\citep{hu2020open}, and 4 new datasets: \dataset{mag-year}, which re-labels \dataset{ogbn-mag} based on publication years to highlight temporal label shifts, and \dataset{oag-cs}, \dataset{oag-eng}, and \dataset{oag-chem}, which are newly constructed from the Open Academic Graph\citep{zhang2019oag}, and reflect disciplinary diversity.

\subsubsection*{\textbf{Social Network Analysis.}}
Social networks provide another example of graphs with both heterophily and heterogeneity. Unlike traditional homophilic assumptions, where friends tend to share similar attributes, real-world social structures reveal connections across diverse demographic and interest groups. Our new \dataset{H-Pokec} dataset, derived from the Pokec social network~\citep{leskovec2016snap}, introduces heterophilic relationships influenced by user demographics and personal affiliations, such as shared hobbies or cultural interests.

\subsubsection*{\textbf{Financial Fraud Detection.}} Fraudulent activities in financial transactions and e-commerce platforms often follow heterophilic patterns: fraudsters attempt to disguise themselves by mimicking normal behaviors with innocent nodes while still forming distinct interaction patterns. Meanwhile, financial networks are inherently heterogeneous, consisting of multiple entity types such as users, businesses, and transactions. Our dataset collection includes a new \dataset{IEEE-CIS-G} graph dataset (developed from a Kaggle tabular dataset \citep{ieee_cis_fraud_2019} for credit card fraud detection in the finance domain) and a repurposed \dataset{RCDD}~\citep{liu2023datasets} dataset (for risk commodity detection in e-commerce domain), both of which capture the heterophilic and heterogeneous nature of financial interactions.

\subsubsection*{\textbf{Malware Detection.}}
Malicious entities on the Internet, such as botnets and phishing domains, do not always form homophilic clusters---they attempt to infiltrate and blend in with legitimate entities. The repurposed \dataset{PDNS} dataset~\citep{kumarasinghe2022pdns} models such behaviors in cybersecurity by representing domain name system (DNS) interactions as a heterogeneous graph, where malicious and benign domains interact with different network entities, making detection a challenging task.

\subsection{Data Standardization}
To ensure consistency, we clean, preprocess, and format all datasets in \benchmark{} following a standardized pipeline. We encapsulate each dataset in the widely used \texttt{HeteroData} object format, supported by the PyTorch Geometric (PyG) library, ensuring seamless compatibility with existing heterogeneous graph learning frameworks. Dataset details are provided in \Cref{{table:stats}} and \Cref{sec:datasets_details}.

\subsubsection{\textbf{Data Formatting and Structure.}}
Each dataset is carefully processed to maintain diverse node/edge types and meaningful graph structures. We ensure that
(1) \textbf{node features are consistently structured}, meaning they share a common representation format across datasets (e.g., numerical embeddings or categorical encodings), facilitating cross-dataset comparisons and model training; and
(2) \textbf{heterogeneous graph information is retained}, with explicit node and edge type definitions stored in the widely used PyG \texttt{HeteroData} format, ensuring compatibility with heterogeneous GNN models.
To facilitate reproducibility and extensibility, new datasets can be integrated into \benchmark{} using our dataset construction script templates, allowing users to format and pre-process data consistently within the framework.

\subsubsection{\textbf{Splitting Strategy.}}
For most datasets, we employ a temporal split scheme, ensuring that the training set precedes the validation set in time, and the validation set precedes the test set. This strategy aligns with real-world prediction scenarios, where models must generalize to future data rather than relying on randomly shuffled samples. Two exceptions are \dataset{mag-year}, where publication year is the prediction target and thus unsuitable for temporal splitting, and \dataset{H-Pokec}, which lacks timestamp information.

\subsection{Data Quantification (\hindex{} Index)}\label{sec:new_heterophily}
To better characterize the structural properties of our datasets, we systematically quantify heterophily in heterogeneous contexts using several standard heterophily metrics and our new metric, the \hindex{} index (\Cref{table:typical_heterophily}).

\begin{figure*}[htb]
    \centering
    \includegraphics[width=0.90\textwidth]{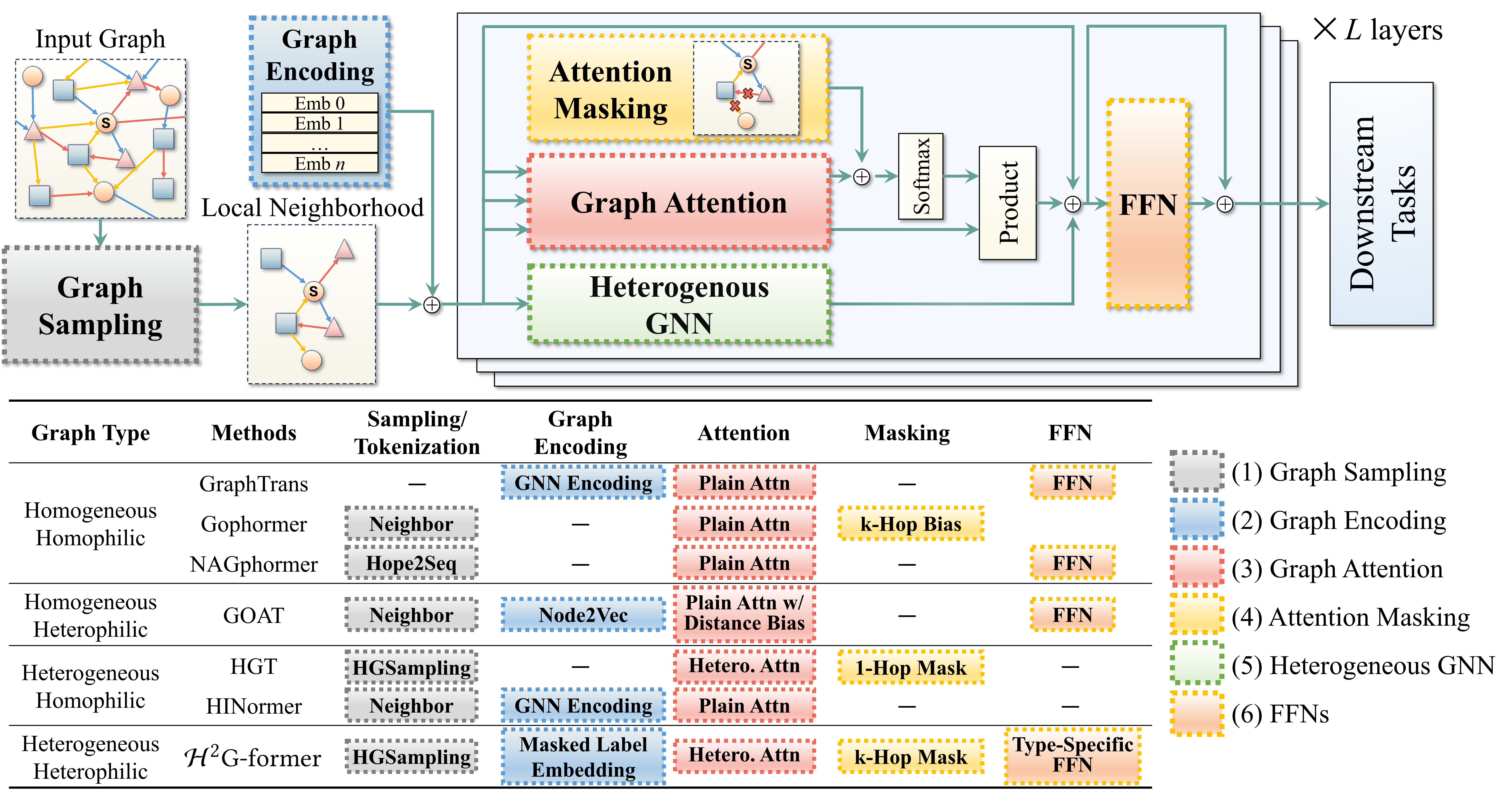}
    \caption{The modular modeling framework (\framework{}) provided by \benchmark{}. We choose several example models from the 28 baselines to demonstrate how they can be reproduced via the modular components provided by the modeling framework.}
    \label{fig:framework}
\end{figure*}

\subsubsection{\textbf{New Heterogeneous Heterophily Metric}}
Inspired by the adjusted heterophily metric~\citep{platonov2024characterizing, suresh2021breaking}, we propose the \textit{class-adjusted heterogeneous heterophily index} \hindex{}, formulated as follows:
\begin{align}
\mathcal{H}^2(\mathcal{G}) &= \text{Agg}\left(\mathcal{H}_{\text{adj}}(\mathcal{G_P}) \left| \mathcal{P}\in \mathcal{M}_k\right.\right),
\end{align}
where $\mathcal{G_P}$ denotes a metapath-induced subgraph, $\mathcal{H}_{\text{adj}}$ is the adjusted heterophily, $\mathcal{M}_k$ is a $k$-hop metapath set, and $\text{Agg} \in \{\text{mean}, \text{max}\}$ is an aggregation function. 
Intuitively, the adjusted heterophily $\mathcal{H}_{\text{adj}}$ quantifies the degree of heterophily relative to what would be expected in a random graph. 
Under the random graph configuration model described in~\cite{molloy1995critical}, where for every node $v$ we create $d(v)$ copies of it and then find a random matching among all nodes, 
the likelihood of a given edge endpoint connecting to a node of class $k$ is approximately $D_k/(2|\mathcal{E}|)$ (as assumed in~\cite{platonov2024characterizing}). Thus, the expected heterophily is the likelihood that two edge endpoints are in different classes, which is
$1-\sum_{k=1}^{C}{D_k(D_k-1)/((2|\mathcal{E}|)(2|\mathcal{E}|-1))} \approx 1-\sum_{k=1}^{C}{D_k^2/(2|\mathcal{E}|)^2}$. 
As a result, a value of \hindex{} close to 0 indicates that nodes predominantly connect to other nodes of the same class, exhibiting homophily. A value approaching or exceeding 1 suggests that nodes are more likely to connect to nodes of different classes, demonstrating heterophily. 
The set of all possible metapaths $\mathcal{P}$ can potentially be large, and so we introduce an additional constraint where only length-2 metapaths are considered. We select the mean function as the aggregation function to reflect the general heterophily across all metapaths. The \hindex{} value for each dataset is presented in \Cref{table:typical_heterophily}.

\subsection{Standard Workflow}
We establish a standard workflow for model developers and application researchers to use \benchmark{}, as shown in \Cref{fig:benchmark}.

\begin{itemize}[topsep=1pt, leftmargin=10pt]
\item \textit{Application Researchers} can search for effective models for their new dataset/application domain as follows:
\begin{enumerate}[topsep=1pt]
\item \textbf{Identify new applications} requiring heterophilic and heterogeneous graph learning.
\item \textbf{Build and integrate dataset} into \benchmark{}.
\item \textbf{Choose models} from our modeling framework.
\item \textbf{Benchmark performance.}
\end{enumerate}

\item \textit{Model Developers} can perform model development as follows:
\begin{enumerate}[topsep=1pt]
\item \textbf{Implement new models} by modifying models in \benchmark{}.
\item \textbf{Evaluate models} to understand performance gaps.
\item \textbf{Iterate} to refine scalable heterophilic and heterogeneous learning approaches.
\end{enumerate}
\end{itemize}

\section{Modular Modeling Framework} 
\label{sec:framework}

\definecolor{mygreen}{RGB}{175, 213, 208}
\definecolor{first}{RGB}{237, 106, 89}
\definecolor{second}{RGB}{255, 192, 0}
\definecolor{third}{RGB}{91, 155, 213}
\newcommand{\accuracycolor}[1]{%
  \cellcolor{mygreen!#1}  %
}
\newcommand{\res}[2]{#1 \textsubscript{$\pm$ #2}}
\newcommand{\fir}[2]{\textcolor{first}{\textbf{#1 \textsubscript{$\pm$ #2}}}}
\newcommand{\sed}[2]{\underline{#1 \textsubscript{$\pm$ #2}}}
\newcommand{\thi}[2]{{#1 \textsubscript{$\pm$ #2}}}
\newcommand{\firi}[1]{\textcolor{first}{\textbf{#1}}}
\newcommand{\seci}[1]{\underline{#1}}
\newcommand{\thii}[1]{{#1}}
\newcommand{\colorres}[3]{\textcolor{#1}{\textbf{#2} \textsubscript{$\pm$ \textbf{#3}}}}

\begin{table*}[t]
\caption{Benchmark results of various GNN methods. Standard deviations are calculated over 5 runs with different random seeds. We highlight the \firi{first} and \seci{second} best results. Label propagation (LP) has deterministic results. Out-of-memory (OOM) indicates the method ran out of memory on an Nvidia V100 GPU with 32GB of memory. \S\S: Heterogeneous Heterophilic.}

\label{table:result}
\resizebox{\textwidth}{!}{%
\begin{tabularx}{1.075\textwidth}{@{}l@{\hspace{0.1em}}l@{\hspace{0.2em}}c@{\hspace{0.8em}}c@{\hspace{0.8em}}c@{\hspace{0.8em}}c@{\hspace{0.8em}}c@{\hspace{0.8em}}c@{\hspace{0.8em}}c@{\hspace{0.8em}}c@{\hspace{0.8em}}c@{\hspace{0.8em}}c@{}}%
\toprule
& \multirow{2}{*}{\thead{\normalsize Datasets$\rightarrow$ \\ \normalsize (\hindex{} Index)}} & \multirow{2}{*}{\thead{\normalsize Avg.\\ \normalsize Rank}} & \multicolumn{6}{c}{Accuracy} & \multicolumn{3}{c}{F1 score}\\
\cmidrule(lr){4-9} \cmidrule(lr){10-12}\textbf{}

&  & & \dataset{ogbn-mag} & \dataset{mag-year} & \dataset{oag-cs} & \dataset{oag-eng} & \dataset{oag-chem} & \dataset{H-Pokec} & \dataset{RCDD} & \dataset{IEEE-CIS-G} & \dataset{PDNS}\\
& Methods$\downarrow$ & & (0.8773) & (0.9654) & (0.9652) & (0.8729) & (0.8858) & (0.9488) & (0.9776) & (0.9846) & (0.7866)\\
\midrule
& MLP & 23.2 & \res{27.27}{0.50} & \res{26.52}{0.64} & \res{09.26}{0.51} & \res{20.18}{0.92} & \res{13.61}{0.41} & \res{62.75}{0.34} & \res{75.87}{1.38} & \res{04.26}{8.52} & \res{73.92}{0.66}\\
\midrule
\multirow{4}{*}{\rotatebox{90}{\thead{Graph Only}}}
& LP+1Hop & 18.9 & 38.36 & 26.61 & 19.79 & 36.07 & 22.48 & 45.42 & 67.07 & 0.00 & 81.53 \\
& LP+2Hop & 14.8 & 37.38 & 39.45 & 20.98 & 36.73 & 21.54 & 76.72 & 67.84 & 0.00 & 82.13 \\
& SGC+1Hop & 24.6 & \res{16.46}{0.24} & \res{26.48}{0.17} & \res{06.42}{0.17} & \res{10.93}{3.18} & \res{07.02}{1.72} & \res{52.91}{0.43} & \res{05.47}{6.92} & \res{13.04}{3.53} & \res{74.24}{1.90}\\
& SGC+2Hop & 25.2 & \res{14.28}{0.28} & \res{26.46}{0.05} & \res{06.09}{0.50} & \res{08.77}{1.22} & \res{05.00}{1.10} & \res{59.55}{1.75} & \res{06.07}{5.29} & \res{07.98}{8.54} & \res{61.34}{1.14}\\
\midrule
\multirow{8}{*}{\rotatebox{90}{\thead{Homogeneous\\Homophilic}}}
& GCN & 14.3 & \res{42.90}{0.50} & \res{32.91}{0.50} & \res{18.22}{0.60} & \res{29.09}{0.52} & \res{18.57}{1.06} & \res{70.63}{0.36} & \res{85.81}{0.87} & \res{28.79}{1.07} & \res{81.22}{0.30}\\
& GraphSAGE & 8.4 & \res{40.80}{0.56} & \res{36.28}{0.19} & \res{22.92}{0.29} & \res{36.16}{0.20} & \res{24.66}{0.48} & \res{77.29}{0.30} & \res{85.02}{0.83} & \thi{31.49}{1.23} & \res{91.44}{0.32} \\
& GAT & 11.7 & \res{48.60}{0.29} & \res{33.50}{0.62} & \res{19.12}{0.25} & \res{28.74}{0.60} & \res{14.05}{0.44} & \res{70.89}{0.20} & \res{86.71}{1.27} & \res{28.51}{0.45} &  \thi{93.97}{0.27} \\
& GIN & 15.6 & \res{37.32}{0.33} & \res{31.15}{0.54} & \res{16.33}{1.34} & \res{29.62}{1.15} & \res{17.86}{0.62} & \res{74.72}{0.32} & \res{84.22}{0.34} & \res{28.53}{0.54} &  \res{87.91}{0.46}\\
& APPNP & 18.3 & \res{37.64}{0.31} & \res{29.79}{0.61} & \res{17.90}{0.60} & \res{28.63}{0.40} & \res{17.19}{1.06} & \res{57.27}{1.22} & \res{82.95}{0.67} & \res{27.27}{1.47} &  \res{80.70}{0.73}\\
& NAGphormer & 11.7 & \res{42.47}{0.74} & \res{32.60}{0.06} & \res{16.49}{0.55} & \res{31.85}{0.80} & \res{23.78}{0.35} & \sed{80.59}{0.15} & \res{85.46}{0.50} & \res{17.07}{0.34} &  \res{92.37}{0.22}\\
& GraphTrans & 13.7 & \res{47.25}{1.54} & \res{36.14}{0.41} & \res{02.39}{0.22} & \res{06.55}{3.53} & \res{02.23}{0.20} & \res{77.80}{0.17} & \res{86.00}{0.56} & \res{30.53}{1.60} & \res{93.00}{0.39}\\
& Gophormer & 16.6 & \res{42.87}{0.64} & \res{35.17}{0.27} & \res{03.68}{1.24} & \res{10.42}{3.73} & \res{04.26}{2.85} & \res{71.55}{2.04} & \res{80.56}{6.13} & \res{30.79}{1.06} & \res{91.58}{0.05}\\
\midrule
\multirow{7}{*}{\rotatebox{90}{\thead{Homogeneous\\Heterophilic}}}
& MixHop & 6.4 & \res{46.99}{0.41} & \res{36.36}{0.28} & \thi{23.04}{0.24} & \res{36.88}{0.73} & \res{25.03}{0.90} & \res{78.78}{0.27} & \res{85.43}{1.22} & \res{30.13}{0.86} &  \res{92.78}{0.18}\\
& LINKX & 12.0 & \res{40.83}{0.18} & \sed{42.81}{0.14} & \res{15.32}{0.08} & \res{32.85}{0.38} & \res{22.98}{0.24} & \thi{79.66}{0.94} & OOM              & \res{31.42}{1.20} &  \res{87.74}{0.52}\\
& FAGCN & 20.9 & \res{33.06}{0.59} & \res{27.10}{0.66} & \res{10.46}{0.44} & \res{22.75}{0.94} & \res{13.01}{0.44} & \res{67.15}{0.09} & \res{81.06}{1.24} & \res{10.09}{5.09} &  \res{82.84}{1.07}\\
& ACM-GCN & 21.1 & \res{33.50}{1.13} & \res{23.20}{1.21} & \res{11.23}{0.75} & \res{22.27}{0.77} & \res{13.81}{0.43} & \res{66.69}{0.09} & \res{75.52}{1.74} & \res{16.98}{0.29} & \res{88.48}{0.48}\\
& LSGNN & 13.6 & \res{38.87}{0.83} & \thi{40.47}{0.58} & \res{15.20}{0.60} & \res{29.43}{0.74} & \res{19.96}{0.69} & \res{78.37}{0.49} & \res{83.84}{0.91} & \res{14.68}{1.86} & \res{88.91}{0.17}\\
& GOAT & 10.3 & \res{41.59}{0.09} & \res{32.92}{0.41} & \res{20.74}{0.39} & \res{35.82}{0.52} & \res{21.75}{0.17} & \res{76.55}{0.71} & \sed{87.13}{0.45} & \res{30.31}{0.73} & \res{91.71}{0.27}\\
& PolyFormer & 17.2 & \res{35.58}{0.24} & \res{31.13}{0.50} & \res{09.22}{0.24} & \res{21.4}{0.62} & \res{15.26}{0.50} & \res{70.74}{0.10} & \res{83.61}{0.69} & \res{17.26}{0.07} & \sed{94.81}{0.09}\\
\midrule
\multirow{7}{*}{\rotatebox{90}{\thead{Heterogeneous\\Homophilic}}}
& R-GCN & \seci{5.3} & \res{46.93}{0.46} & \res{35.60}{0.48} & \sed{23.10}{1.09} & \sed{37.10}{0.49} & \res{25.80}{0.32} & \res{78.05}{0.28} & \thi{87.00}{1.35} & \res{31.44}{0.96} &  \res{92.55}{0.44}\\
& \small R-GraphSAGE & 6.0 & \sed{50.94}{0.44} & \res{38.07}{0.41} & \res{22.81}{0.63} & \res{36.11}{0.45} & \thi{26.00}{0.59} & \res{77.00}{0.32} & \res{86.81}{1.74} & \res{29.85}{0.47} & \res{92.81}{0.37}\\
& R-GAT & 11.0 & \res{41.51}{0.47} & \res{35.40}{0.88} & \res{21.03}{0.59} & \res{35.90}{0.60} & \sed{26.14}{0.34} & \res{67.17}{0.24} & \res{80.37}{0.62} & \res{22.09}{0.94} &  \res{94.29}{0.16}\\
& HAN & 19.1 & \res{39.00}{0.22} & \res{29.66}{0.43} & \res{13.14}{1.96} & \res{27.81}{0.69} & \res{17.03}{0.66} & \res{54.04}{2.17} & \res{78.56}{1.42} & \res{23.15}{0.43} &  \res{84.58}{0.76}\\
& HGT & \thii{5.9} & \thi{50.23}{0.48} & \res{39.47}{1.66} & \res{22.51}{0.40} & \res{35.51}{0.52} & \res{25.48}{0.76} & \res{78.91}{0.43} & \res{86.05}{1.01} & \res{30.89}{0.80} &  \res{92.76}{0.15} \\
& HINormer & 27.7 & OOM & OOM & OOM & OOM & OOM & OOM & OOM & OOM & OOM\\
& SHGN & 11.0 & \res{43.39}{0.28} & \res{34.43}{1.23} & \res{22.03}{0.46} & \thi{36.93}{0.67} & \res{24.07}{0.94} & \res{50.50}{0.89} & \res{79.67}{2.53} & \fir{31.66}{0.86} & \res{89.33}{0.21}\\
\midrule
\thead{\large\S\S} & \model & \firi{1.1} & \fir{55.67}{0.35} & \fir{52.55}{0.66} & \fir{28.47}{0.93} & \fir{46.63}{0.65} & \fir{30.62}{0.31} & \fir{82.45}{0.19} & \fir{87.35}{0.80} & \sed{31.55}{0.92} & \fir{96.43}{0.21}\\
\bottomrule
\end{tabularx}
}
\end{table*}

To facilitate standardized benchmarking, \benchmark{} incorporates \framework{} \citep{lin2024unifiedgt}, a modular modeling framework that we previously designed that is capable of expressing various GNN architectures, as shown in \Cref{fig:framework}. \framework{} provides a structured approach to decomposing graph learning models into modular components, including graph sampling, encoding, attention mechanisms, heterogeneous GNN, and feedforward networks (FFN), allowing flexible integration of different modeling techniques.

The modeling framework enables flexible experiments and performance comparisons across 28 state-of-the-art baseline models, reducing implementation variability and simplifying the process of integrating new models into \benchmark{}. The modeling framework provides simple baselines and three categories of state-of-the-art GNN and graph transformer models. The simple baselines include models that only consider node features, such as MLP~\citep{goodfellow2016deep}, and models that only consider graph topology, such as label propagation (LP, one and two hops)~\citep{zhou2003learning, peel2017graph}, as well as a simple GNN model that focuses on aggregation of neighborhood information with reduced nonlinearities and weight matrices, SGC~\citep{wu2019simplifying}. The first class of GNN baselines, designed for \textit{homogeneous homophilic} graphs, includes GCN~\citep{kipf2016semi}, GraphSAGE~\citep{hamilton2017inductive}, GAT~\citep{velickovic2017graph}, GIN~\citep{xu2018powerful}, APPNP~\citep{gasteiger2019predict}, NAGphormer~\citep{chen2022nagphormer}, GraphTrans~\citep{wu2021representing}, and Gophormer~\citep{zhao2021gophormer}. 
The second class of baselines, optimized for \textit{homogeneous heterophilic} graphs, includes MixHop~\citep{abu2019mixhop}, LINKX~\citep{lim2021large}, FAGCN~\citep{bo2021beyond}, ACM-GCN~\citep{luan2022revisiting}, LSGNN~\citep{chen2023lsgnn}, GOAT~\citep{kong2023goat}, and PolyFormer~\citep{ma2024polyformer}.
The third class of baselines, designed for \textit{heterogeneous homophilic} graphs, includes relational GCN (R-GCN)~\citep{schlichtkrull2018modeling}, GraphSAGE (R-GraphSAGE), GAT (R-GAT), HAN~\citep{wang2019heterogeneous}, HGT~\citep{hu2020heterogeneous}, HINormer~\citep{mao2023hinormer}, and SHGN~\citep{lv2021we}.
Lastly, we present a new model \model{} designed for \textit{heterogeneous heterophilic} graphs, developed following our established workflow (\Cref{sec:case_study}).
The detailed descriptions of each model can be found in \Cref{sec:baselines}.

\section{Experiments}
\label{sec:experiments}
In this section, we conduct comprehensive experiments to evaluate existing and proposed methods in \benchmark{} using an Nvidia V100 GPU with 32GB of memory. The homogeneous methods ignore the node and edge types.

\subsection{General Setup}

\subsubsection{\textbf{Training and Evaluation.}} The dataset splits can be found at \Cref{table:stats}, where most of the split strategy is based on timestamps on the nodes.
Test performance is reported for the learned parameters corresponding to the highest validation performance. We use F1 score as the metric for the datasets with large class imbalance, as it is less sensitive to class imbalance than accuracy. For the other datasets, we use classification accuracy as the metric.

\subsubsection{\textbf{Minibatching Sampling.}} Most existing heterophilic GNNs are designed for small graphs and struggle to scale to large graphs. To enable training on large graphs, our framework supports optional minibatching, where models process sampled local neighborhoods instead of the full graph. In our experiments, we adopt minibatching for scalability, using a consistent sampling strategy across all models within each dataset to ensure fair comparison.
While some models may benefit from specialized sampling, varying strategies would introduce confounding factors that obscure model-level effects.

\begin{figure}[t]
    \centering
    \includegraphics[width=0.95\columnwidth]{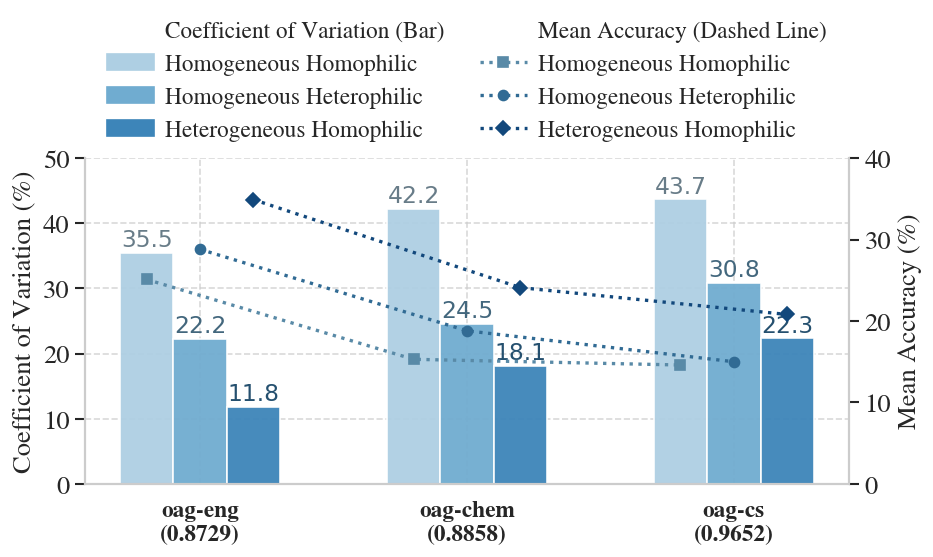}
    \caption{Model group performance versus heterophily. The coefficient of variation is the standard deviation of model accuracy in each group normalized by the mean accuracy. The \hindex{} index is indicated under each dataset name. }
    \label{fig:variations}
\end{figure}

\subsection{Experimental Results}
\Cref{table:result} lists the results of each method across the datasets proposed in \benchmark. We make the following observations:
\begin{enumerate}[topsep=1pt, leftmargin=12pt]
\item[(1)] \textbf{\model{} consistently outperforms baselines across diverse graph structures.} It achieves the best average rank (1.1) and consistently outperforms or matches the existing methods on all of the datasets. This highlights its ability to effectively capture both heterophilic and heterogeneous structures, reinforcing the need for models tailored to such real-world graphs.
\item[(2)] \textbf{Homogeneous heterophilic GNNs struggle with heterogeneous graphs.} While methods like MixHop and GOAT outperform homogeneous homophilic GNNs in our benchmark, achieving a better average rank, their advantage diminishes when compared to heterogeneous homophilic GNNs. This performance degradation primarily stems from their inability to effectively incorporate diverse node and edge types. For example, the semantic meaning of each type of node can be different, resulting in different distributions in the node features. These homogeneous heterophilic GNNs cannot adjust their parameters to learn from node features of different distributions. 
\item[(3)] \textbf{Performance of heterogeneous homophilic GNNs depends on their ability to handle heterophily.} The performance of heterogeneous models varies significantly, likely due to differences in their architectural robustness when exposed to heterophily. For instance, models relying on local attention mechanisms (e.g., R-GAT, HAN, and SHGN compute attention over 1-hop neighbors) generally underperform. We quantitatively illustrate this in \Cref{fig:variations}, where we select three datasets from a single domain (academic networks), with similar heterogeneity (number of nodes/edge types) but different heterophily. We evaluate the performance variations within each model group, and can clearly observe that datasets with higher heterophily (e.g., \dataset{oag-cs}) show greater variations across models within the group. Consistent with observation (2), we also observe that heterogeneous models perform better, with lower variations and higher mean accuracy, emphasizing the importance of effectively handling the different node and edge types in achieving good task performance. Building on this insight, our \model{} incorporates $k$-hop attention, instead of 1-hop attention, and considers the graph heterogeneity, leading to improved performance.

\item[(4)] \textbf{Scalability issues in existing GNNs.} 
A significant gap exists between the best and worst-performing homogeneous heterophilic GNNs, particularly as the graph size increases. Many of these GNNs were designed for small-scale datasets and full-graph training and struggle when trained on large-scale graphs using mini-batching. For example, FAGCN and ACM-GCN show degraded performance, consistent with observations in the previous work~\citep{lim2021large}. This underscores the need for scalable architectures that can handle both heterophily and heterogeneity.

\item[(5)] \textbf{Dataset-specific insights: how performance varies by domain.}
Our results demonstrate that certain model types perform well in specific domains but fail in others, emphasizing the importance of a diverse benchmark. 
In academic networks (e.g., \dataset{ogbn-mag} and \dataset{oag-cs}), R-GraphSAGE and R-GCN perform well, leveraging hierarchical information from paper-author-affiliation relationships. Homogeneous heterophilic models struggle, as they lack relational reasoning over entity types.
In e-commerce and security networks (e.g., \dataset{RCDD} and \dataset{PDNS}), GOAT and PolyFormer perform well, suggesting that effective handling of long-range dependencies and robust graph structure encoding are crucial in fraud and security applications. 
In social networks (e.g., \dataset{H-Pokec}), the homophilic model NAGphormer performs surprisingly well, likely due to its ability to aggregate information from multi-hop neighborhoods, effectively capturing long-range homophilic signals. We also observe that models leveraging heterophilic signals, such as MixHop and LSGNN, achieve relatively strong performance by addressing heterophily among labeled users. However, they still underperform compared to \model{}, as they fail to exploit the rich metapath information embedded in the graph.

\end{enumerate}

\begin{figure}[t]
    \centering
    \includegraphics[width=0.95\columnwidth]{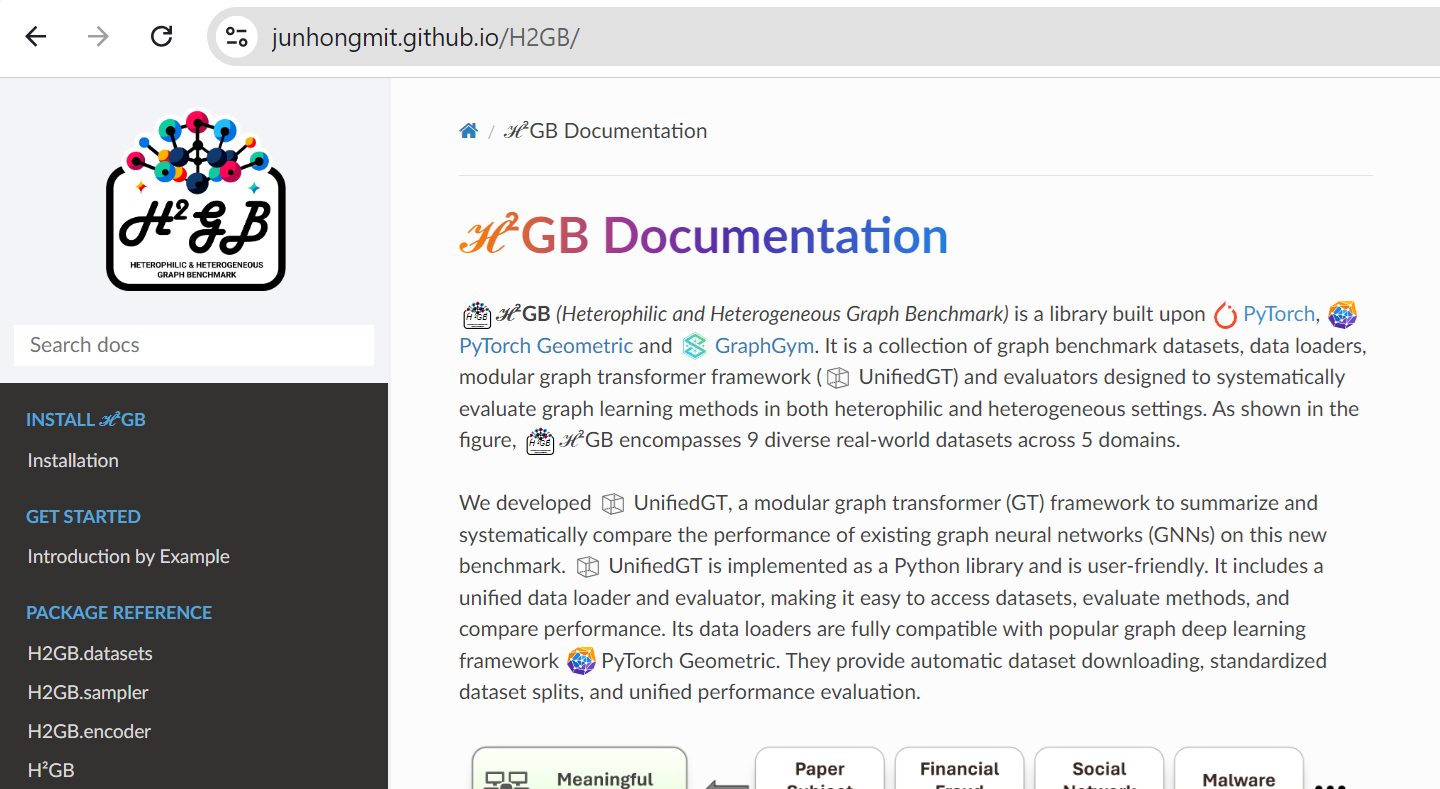}
    \caption{\benchmark{} has a user-friendly website and provides an introduction with examples.}
    \label{fig:website}
\end{figure}

\subsection{Case Study: \benchmark{} for Model Development}\label{sec:case_study}
\benchmark{} provides user-friendly examples (\Cref{fig:website}) and facilitates research following our standardized workflow. We present a case study on the construction of the \dataset{oag-cs} dataset and the development of the \model{} model.
\begin{itemize}[topsep=1pt, leftmargin=10pt]
\item \noindent \textbf{Step 1: Identifying Application.} We aim to predict which venue a computer science paper will be published in, a challenging task due to the diverse paper-author-affiliation interactions and interdisciplinary nature of research.

\item \textbf{Step 2: Building and Standardizing Dataset.} Using the Open Academic Graph (OAG), we extract papers in the computer science field to construct an academic network. We represent node features using paper abstract embeddings and define multiple node types, including papers, authors, affiliations, and topics, along with their interactions as edge types. Publication venues serve as node labels. This dataset is integrated into \benchmark{} as \dataset{oag-cs} and made accessible through our standardized data loader.

\item \textbf{Step 3: Evaluating Baselines and Identifying Limitations.} We evaluated all baselines and found the best accuracy to be 23.10\%, meaning that fewer than a quarter of papers are correctly classified. This suggests room for improvement.

\item \textbf{Step 4: Iterative Model Development using Modular Components.} To demonstrate how our benchmark can facilitate principled model design, we use \framework{} to systematically enhance a strong baseline, HGT (22.51\%). As shown in \Cref{fig:framework}, HGT consists of: \textit{HGSampling} (Graph Sampling), \textit{Heterogeneous Attention} (Graph Attention), and \textit{1-Hop Mask} (Attention Masking). We experiment with component-level modifications: replacing the 1-Hop Mask with \textit{k-Hop Mask} (enabling better context capture), enhancing graph encoding with \textit{masked label embeddings} (which assists in predicting node labels), and introducing a \textit{Type-Specific FFN} since HGT lacks a dedicated FFN before the output. This modular modification process results in our new method, \model{}, illustrating how \benchmark{} enables targeted model development through interpretable architecture changes.

\item \textbf{Step 5: Results.}
With these modifications, the accuracy improves to 28.47\% shown in \Cref{table:result}, a 5.37\% improvement over the best baseline. This demonstrates how \benchmark{} enables systematic model evaluation and component-wise experimentation, making it a powerful toolbox for benchmarking and research.
\end{itemize}

\section{Conclusion}
\label{sec:conclusion}
We introduce \benchmark{}, a comprehensive benchmark for evaluating graph learning models on large-scale real-world heterophilic and heterogeneous graphs. We provide a unified benchmarking library with a standardized data loader, evaluator, and extensible framework for systematic experimentation. Our comprehensive benchmarking on 28 baseline models highlights the challenges posed by heterophilic and heterogeneous graphs and provides insights into model performance. Through a case study, we demonstrate how \benchmark{} facilitates model selection and guides the development of improved methods such as \model{}. 
We believe \benchmark{} serves as a vital resource for advancing scalable and realistic graph learning research.
Directions for future work include incorporating more datasets into \benchmark{} and extending datasets and models to other tasks such as link prediction and node regression.

\begin{acks} 
This work is funded by the MIT-IBM AI Watson Lab, NSF awards \#CCF-1845763, \#CCF-2316235, and \#CCF-2403237, Google Faculty Research Award, and Google Research Scholar Award. We thank Dawei Zhou (Virginia Tech) for his valuable feedback and guidance.
\end{acks}

\bibliographystyle{ACM-Reference-Format}
\bibliography{reference}

\appendix
\section{Dataset Documentation, Metadata, and Intended Use}
All datasets in \benchmark{} are intended for academic use, and their corresponding licenses are described in \Cref{sec:license}. We release our \benchmark{} as an open-source library under the \underline{MIT license}. 
For ease of access, we provide the following links to the \benchmark{} benchmark suite and \framework{} framework:
\begin{itemize}[topsep=1pt,itemsep=2pt,parsep=0pt,leftmargin=10pt]
\item The open-source library is at {\textcolor{blue}{\url{https://github.com/junhongmit/H2GB/}}}.
\item The \benchmark{} \texttt{Python} package is at \textcolor{blue}{\url{https://pypi.org/project/H2GB}}.
\item Datasets and documentation are at {\textcolor{blue}{\url{https://junhongmit.github.io/H2GB/}}}.
\end{itemize}

\paragraph{\textbf{Croissant Metadata.}} Croissant metadata records documenting each dataset can be found at
\begin{itemize}[topsep=1pt,itemsep=2pt,parsep=0pt,leftmargin=10pt]
\item \dataset{ogbn-mag}, \dataset{mag-year}: \textcolor{blue}{\href{https://www.kaggle.com/datasets/junhonglin/mag-year-dataset-for-h2gb/croissant/download}{Croissant metadata}}.
\item \dataset{oag-cs}, \dataset{oag-eng}, \dataset{oag-chem}: \textcolor{blue}{\href{https://www.kaggle.com/datasets/junhonglin/oag-dataset-for-h2gb/croissant/download}{Croissant metadata}}.
\item \dataset{RCDD:} \textcolor{blue}{\href{https://www.kaggle.com/datasets/junhonglin/rcdd-dataset-for-h2gb/croissant/download}{Croissant metadata}}.
\item \dataset{IEEE-CIS}: \textcolor{blue}{\href{https://www.kaggle.com/datasets/junhonglin/ieee-cis-dataset-for-h2gb/croissant/download}{Croissant metadata}}.
\item \dataset{H-Pokec:} \textcolor{blue}{\href{https://www.kaggle.com/datasets/junhonglin/h-pokec-dataset-for-h2gb/croissant/download}{Croissant metadata}}.
\item \dataset{PDNS}: \textcolor{blue}{\href{https://www.kaggle.com/datasets/junhonglin/pdns-dataset-for-h2gb/croissant/download}{Croissant metadata}}.
\end{itemize}

\section{Additional Dataset Details} 
\subsection{Licenses}\label{sec:license}
In this section, we indicate the licenses of the collected datasets:
\begin{itemize}[topsep=1pt,itemsep=2pt,parsep=0pt,leftmargin=10pt]
\item {\dataset{ogbn-mag}, \dataset{mag-year}, \dataset{oag-cs}, \dataset{oag-eng}, \dataset{oag-chem}}: \underline{\textbf{ODC-BY}}. Licensed via Open Graph Benchmark~\citep{hu2020open} and Open Academic Graph~\citep{zhang2019oag}.
\item \dataset{RCDD}: \underline{\textbf{CC BY 4.0}}. Publicly released~\citep{liu2023datasets}. Node/edge type names are redacted for confidentiality; features are numeric.
\item \dataset{IEEE-CIS}: Released via the IEEE CIS Kaggle challenge~\citep{ieee_cis_fraud_2019}, with anonymized transaction records and numeric-only features.  To the best of our knowledge, it was not released with a license.
\item \dataset{Pokec}: \underline{\textbf{BSD}}. Provided via SNAP~\citep{takac2012data, leskovec2016snap}. Text features are removed; only numeric features are retained for privacy.
\item \dataset{PDNS}: Publicly released~\citep{kumarasinghe2022pdns}, with anonymized graphs and numeric-only features. To the best of our knowledge, the dataset was not released with a license.
\end{itemize}

\subsection{Dataset Details.}\label{sec:datasets_details}
All datasets in \benchmark{} are formatted as \texttt{HeteroData} objects compatible with PyTorch Geometric. We summarize each dataset below.
\begin{itemize}[topsep=1pt,itemsep=2pt,parsep=0pt,leftmargin=10pt]
\item \dataset{ogbn-mag}~\citep{hu2020open}: A heterogeneous academic graph with papers, authors, institutions, and fields of study, connected via four relation types. Paper nodes have 128-dimensional Word2Vec~\citep{mikolov2013efficient} features; others are initialized via mean aggregation. Labels denote paper venues. We adopt the official temporal split: training (pre-2018), validation (2018), testing (post-2018).

\item \dataset{mag-year}~\citep{hu2020open}: Same structure as \dataset{ogbn-mag}, but paper labels correspond to publication year buckets (5 balanced classes).

\item \dataset{oag-cs}, \dataset{oag-eng}, and \dataset{oag-chem}~\citep{zhang2019oag}: Subsets of OAG for computer science, engineering, and chemistry, respectively. Entities and relations match \dataset{ogbn-mag}. Paper nodes use 768-dim XLNet~\citep{yang2019xlnet} embeddings of their titles. Labels are paper venues. We apply a temporal split: train (pre-2017), val (2017), test (post-2017).

\item \dataset{RCCD} (Risk Commodity Detection Dataset)~\citep{liu2023datasets}: A large-scale heterogeneous e-commerce graph from Alibaba. Node/edge types (except for items) are anonymized. Item nodes have 256-dimemsional features (BERT~\citep{devlin2018bert} + BYOL~\citep{grill2020bootstrap}). Labels indicate risk commodities (binary). We follow the official split, where the validation set is split from the training set, and the test set is obtained over time.

\item \dataset{IEEE-CIS-G}~\citep{ieee_cis_fraud_2019}: A bipartite financial graph from a Kaggle fraud detection dataset. Nodes include transactions and 11 types of transaction metadata (e.g., card info, email domains). Edges link transactions to metadata (22 relation types). Each transaction has a 4823-dimensional feature vector. Fraud labels are binary; ~4\% are positive. A temporal split is used for evaluation.

\item \dataset{H-Pokec}~\citep{takac2012data}: A social network graph with users and hobby club entities. Edges capture friendships and affiliations. User nodes have 66-dimensional profile-based features and gender labels. We apply a random split.

\item \dataset{P-DNS}~\citep{kumarasinghe2022pdns}: A cybersecurity graph of domain and IP nodes from passive DNS logs. Edges include resolutions and domain similarity. Domain nodes have 10-dimensional features (e.g., subdomain count, impersonation flags) and binary labels for maliciousness. We use a temporal split based on resolution time.
\end{itemize}

\begin{figure*}[t!]
    \centering
    \includegraphics[width=0.85\textwidth]{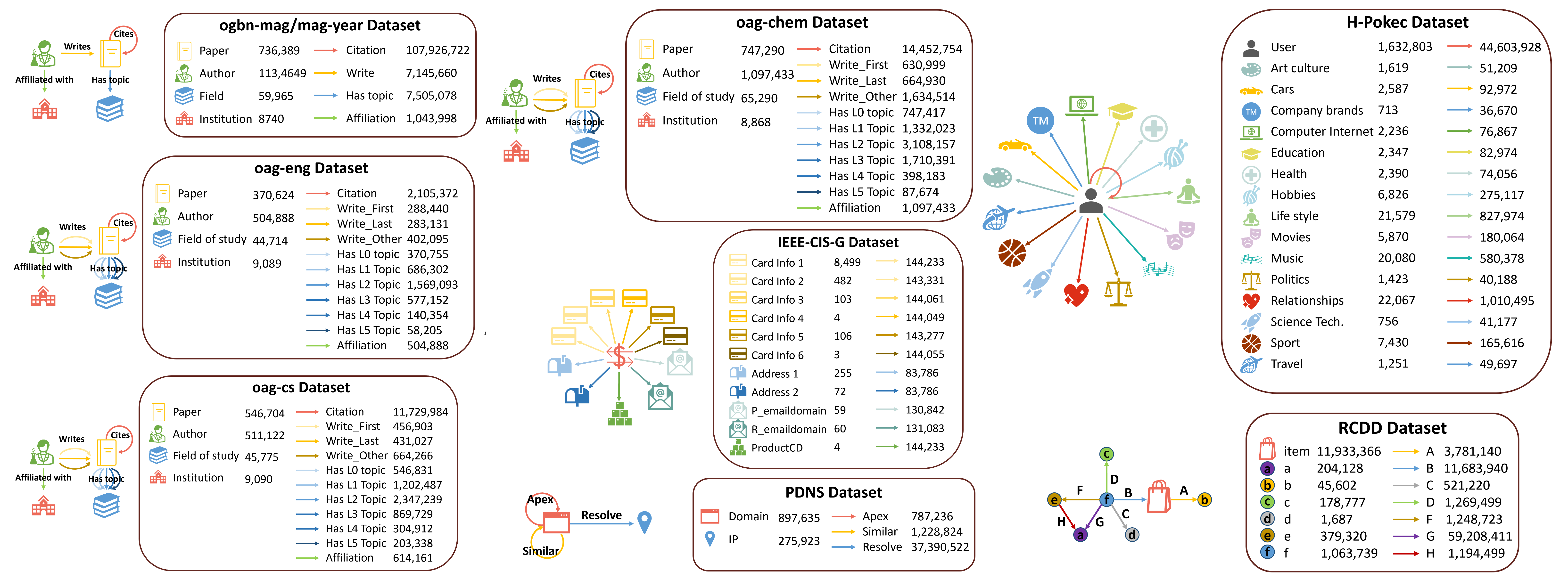}
    \caption{\small The schema and node/edge information of each dataset in \benchmark{}.}
    \label{fig:schema}
\end{figure*}

\Cref{fig:schema} illustrates the heterogeneous graph schema for each dataset. Each schema is a type-level graph, where nodes represent node types and edges denote relation types. Legends indicate the number of nodes and edges per type.

\section{Experiment Setup} \label{sec:details} 
Experiments are implemented in Python 3.9 using PyTorch 2.0.1~\citep{paszke2019pytorch} (\underline{\textbf{BSD-3 license}}) and PyTorch Geometric 2.5.0~\citep{fey2019fast} (\underline{\textbf{MIT license}}). \framework{} builds on GraphGym~\citep{you2020design} (\underline{\textbf{MIT license}}), offering modular components and flexible configuration. We provide experiment configurations for full reproducibility. All training and preprocessing were conducted on an Nvidia V100 GPU (32GB memory).

\subsection{Additional Details of Baselines}
\label{sec:baselines}
Baselines include five groups:
(1) node-only methods, (2) structure-only methods, (3) homogeneous homophilic GNNs, (4) homogeneous heterophilic GNNs, and (5) heterogeneous homophilic GNNs.

\begin{enumerate}[topsep=1pt,itemsep=2pt,parsep=0pt,leftmargin=15pt]
\item[(1)] \textbf{\textit{Node-only.}} \textbf{MLP}~\citep{goodfellow2016deep} ignores the graph structure.
\item[(2)] \textbf{\textit{Structure-only.}} 
\textbf{Label propagation}~\citep{zhou2003learning, peel2017graph}: Spreads labels based on graph connectivity.
\textbf{SGC}~\citep{wu2019simplifying}: Linearizes GCN by collapsing weight matrices and removing nonlinearities.
\item[(3)] \textbf{\textit{Homogeneous Homophilic GNNs.}} 
\textbf{GCN}~\citep{kipf2016semi}: 
A GNN that uses a localized first-order approximation of spectral graph convolutions.
\textbf{GraphSAGE}~\citep{hamilton2017inductive}: A GNN that employs a sampling and aggregation framework to efficiently generate node embeddings. It concatenates the self-node features with neighbors' features and has been shown to perform well when the graph exhibits some heterophily~\citep{zhu2020beyond}.
\textbf{GAT}~\citep{velickovic2017graph}: A GNN that employs the attention mechanism to weight the significance of neighbors.
\textbf{GIN}~\citep{xu2018powerful}: A GNN designed to capture the power of the Weisfeiler-Lehman graph isomorphism test by using a sum aggregator to update the node representations.
\textbf{APPNP}~\citep{gasteiger2019predict}: A GNN that combines the propagation of labels throughout a graph with a personalized PageRank scheme for effective learning.
\textbf{NAGphormer}~\citep{chen2022nagphormer}: A transformer-based GNN that integrates node features and graph topology through attention mechanisms.

\item [(4)] \textbf{\textit{Homogeneous Heterophilic GNNs.}} 
\textbf{MixHop}~\citep{abu2019mixhop}: A heterophilic GNN that aggregates features from a node’s neighbors at various distances, allowing the model to learn more complex patterns of heterophily.
\textbf{FAGCN}~\citep{bo2021beyond}: A heterophilic GNN with improved aggregation mechanisms considering the influence of neighboring nodes based on their label discrepancy.
\textbf{ACM-GCN}~\citep{luan2022revisiting}: A heterophilic GNN designed to discriminate between different types of node relationships.
\textbf{LINKX}~\citep{lim2021large}: A heterophilic GNN that decouples structure and feature transformation, making it simple and scalable.
\textbf{LSGNN}~\citep{chen2023lsgnn}: A heterophilic GNN that models heterophily using local similarity and has been shown to outperform powerful heterophilic GNNs, such as GloGNN~\citep{li2022finding}.

\item [(5)] \textbf{\textit{Heterogeneous Homophilic GNNs.}} 
\textbf{RGCN}~\citep{schlichtkrull2018modeling}: A heterogeneous GNN that introduces relation-specific transformations to separately aggregate neighbors based on relations.
\textbf{RGraphSAGE}: GraphSAGE extended to handle heterogeneous graphs by incorporating edge-type information into the aggregation process.
\textbf{RGAT}: GAT extended to heterogeneous graphs by integrating relational attention into its computation.
\textbf{HAN}~\citep{wang2019heterogeneous}: A GNN that applies both node-level and semantic-level attention, focusing on information aggregation along different metapaths.
\textbf{HGT}~\citep{hu2020heterogeneous}: A heterogeneous GNN that introduces a type-aware attention mechanism to learn node and edge type-dependent representations.
\textbf{SHGN}~\citep{lv2021we}: A heterogeneous GNN that improves node representation learning by leveraging type-specific embeddings, incorporating attention mechanisms and residual connections, and applying an $\ell_2$-norm to the output for regularization and stability.
\textbf{HINormer}~\citep{mao2023hinormer}: A heterogeneous GNN that uses a long-range aggregation mechanism for node representation learning by using a local structure encoder and a heterogeneous relation encoder.

\end{enumerate}

\subsection{Implementation Details} \label{sec:implementation}
\begin{enumerate}[topsep=1pt,itemsep=2pt,parsep=0pt,leftmargin=15pt]
\item \textit{\textbf{Experiment Configurations.}} Hyperparameters are initialized based on official settings and tuned for each dataset. All configurations are available at \textcolor{blue}{\url{https://github.com/junhongmit/H2GB/}}.

\item \textbf{\textit{Minibatching.}} Many heterophilic GNNs are not scalable to large graphs. We apply minibatching using fixed sampling parameters across models to avoid OOM errors and ensure fair comparisons.

\item\textbf{\textit{Graph Encoding.}} \textbf{Featureless Nodes:} Learnable embeddings are assigned to node types lacking input features, such as in \dataset{H-Pokec} and \dataset{IEEE-CIS}.
\textbf{Feature Projection:} All features are projected into a shared embedding space.

\item \textbf{\textit{Model Adaptation.}}
\textbf{Relational Extensions:} We adapt GraphSAGE and GAT to heterogeneous graphs via PyG’s relational wrappers, creating R-GraphSAGE and R-GAT.
\textbf{Optimized Attention:} We provide an efficient cross-type heterogeneous attention implementation using sparse operations to handle fragmented edge representations in PyG/DGL.

\item \textbf{\textit{Optimization.}} We use the AdamW optimizer with cosine annealing and warmup~\citep{loshchilov2016sgdr}, with weight decay set to $10^{-5}$.

\end{enumerate}

\end{document}